\documentclass[runningheads]{llncs}
\usepackage{graphicx}
\usepackage{booktabs} 
\usepackage{multirow} 
\usepackage{bbm} 
\usepackage{amssymb}
\usepackage{wrapfig} 
\usepackage{array}

\usepackage[marginal]{footmisc}

\usepackage{bm}
\usepackage{amsmath}
\usepackage{hyperref}

\usepackage{pifont} 
\usepackage{mathtools} 

\usepackage{graphicx}
\usepackage{caption}
\usepackage{subcaption}

\usepackage{algorithm} \usepackage{algpseudocode}

\usepackage{color}
\usepackage{bbding}

\begin{document}
\setlength{\textfloatsep}{10pt}
\title{PANS: Probabilistic Airway Navigation System for Real-time Robust Bronchoscope Localization}
%
\titlerunning{Probabilistic Airway Navigation System for Bronchoscope Localization}
%
\author{Qingyao Tian\inst{1,2} \and
Zhen Chen\inst{5} \and
Huai Liao\inst{3}\and
Xinyan Huang\inst{3}\and
Bingyu Yang\inst{1,2}\and
Lujie Li\inst{4}\and
Hongbin Liu\inst{1,5}\Envelope
}
%
\authorrunning{Q. Tian et al.}
%
\institute{Institute of Automation, Chinese Academy of Sciences, Beijing, China
\email{liuhongbin@ia.ac.cn}\and
School of Artificial Intelligence, University of Chinese Academy of Sciences, Beijing, China\and
Dept. of Pulmonary and Critical Care Medicine, The First Affiliated Hospital of Sun Yat-sen University, Guangzhou, China\and
Dept. of Radiology, The First Affiliated Hospital, Sun Yat-sen University, Guangzhou, China\and
Centre for Artificial Intelligence and Robotics, Chinese Academy of Sciences, Hong Kong, China}
%
\maketitle              
%
\begin{abstract}
Accurate bronchoscope localization is essential for pulmonary interventions, by providing six degrees of freedom (DOF) in airway navigation. However, the robustness of current vision-based methods is often compromised in clinical practice, and they struggle to perform in real-time and to generalize across cases unseen during training. To overcome these challenges, we propose a novel Probabilistic Airway Navigation System (PANS), leveraging Monte-Carlo method with pose hypotheses and likelihoods to achieve robust and real-time bronchoscope localization. Specifically, our PANS incorporates diverse visual representations (\textit{e.g.}, odometry and landmarks) by leveraging two key modules, including the Depth-based Motion Inference (DMI) and the Bronchial Semantic Analysis (BSA). To generate the pose hypotheses of bronchoscope for PANS, we devise the DMI to accurately propagate the estimation of pose hypotheses over time. Moreover, to estimate the accurate pose likelihood, we devise the BSA module by effectively distinguishing between similar bronchial regions in endoscopic images, along with a novel metric to assess the congruence between estimated depth maps and the segmented airway structure. Under this probabilistic formulation, our PANS is capable of achieving the 6-DOF bronchoscope localization with superior accuracy and robustness. Extensive experiments on the collected pulmonary intervention dataset comprising 10 clinical cases confirm the advantage of our PANS over state-of-the-arts, in terms of both robustness and generalization in localizing deeper airway branches and the efficiency of real-time inference. The proposed PANS reveals its potential to be a reliable tool in the operating room, promising to enhance the quality and safety of pulmonary interventions.

\end{abstract}
\section{Introduction}
Bronchoscopy has been a fundamental tool for examining and diagnosing airway lesions, as well as performing biopsies for lung disorders \cite{andolfi2016role}. During bronchoscopy, surgeons utilize a camera-equipped flexible endoscope to perform a thorough inspection of the accessible bronchial branches and to navigate to lung peripheries and nodules as seen in pre-operative CT scans. However, the bronchoscope's limited field of view demands significant clinical experience for accurate localization within the airways. Therefore, there is an urgent demand for automatic methods to localize the bronchoscope with 6 DOF to assist airway navigation.

Various technologies have emerged to address this challenge, including electromagnetic (EM) navigation \cite{folch2020sensitivity}, 3D shape sensing \cite{floris2021fiber}, and visually navigated bronchoscopy (VNB). Particularly, VNB emerges as a promising area of study for its potential of accurate localization with cost-effectiveness. Existing VNB approaches predominantly utilize singular visual cues for bronchoscope tracking, which are achieved by registration \cite{mori2002tracking,deguchi2009selective,shen2019context,banach2021visually,sganga2019offsetnet}, retrieval \cite{zhao2019generative,sganga2019autonomous}, or visual odometry \cite{deng2023feature,borrego2023bronchopose}. Despite these efforts, current VNB studies are still in a developmental phase, facing several barriers to clinical implementation. Validation relies on biased and limited datasets that mainly cover proximal branches, such as insertion to a target site \cite{sganga2019autonomous,banach2021visually,tian2024ddvnb}, casting doubts on their applicability in diverse clinical scenarios. Additionally, the absence of real-time operation in many VNB solutions \cite{shen2019context,wang2020visual,banach2021visually,gu2022vision,luo2023monocular} and the necessity for extensive, case-specific training \cite{sganga2019autonomous,zhao2019generative} hinder their integration into routine clinical workflows. Consequently, VNB techniques are constrained by their limited robustness, absence of real-time performance, and lack of adaptability. 


The challenge of limited robustness in VNB suggests the adoption of strategies used by experienced surgeons, who employ a comprehensive visual approach, including monitoring bronchoscopic motion and identifying anatomical landmarks, to pinpoint the bronchoscope's location. In response, we present the Probabilistic Airway Navigation System (PANS), enhancing localization robustness through a Monte Carlo framework. Replicating the surgeons' approach, PANS integrates diverse visual representations (\textit{e.g.}, odometry and landmarks) for bronchoscope tracking. Specifically, PANS incorporates the Depth-based Motion Inference (DMI) module to propagate endoscopic pose hypotheses overtime, and the Bronchial Semantic Analysis (BSA) module that enhances pose likelihood measurement by distinguishing airway landmarks in bronchoscopic frames. The DMI module enhances case generalization by initially estimating the depth of incoming endoscopic frames prior to inferring camera motion. This transformation from RGB frames to the depth domain mitigates the issue posed by patient-specific textures and variable endoscopic illumination conditions, enabling a motion inference network trained on synthetic data to adapt to real endoscopic images effectively. By distinguishing airway landmarks, the BSA module adeptly clarifies ambiguities in similar airway areas, thereby significantly enhancing the overall precision of localization. It employs a novel metric for evaluating the congruence between depth estimations and segmented airway models, enhancing pose estimation's robustness. Furthermore, for real-time application across varied cases, PANS incorporates optimized lightweight neural networks for depth estimation, motion inference, and landmark detection. We perform an extensive comparative evaluation against SOTA methods, showcasing the superior performance of our PANS on real clinical datasets. Additionally, thorough ablation studies highlight the effectiveness of leveraging multiple visual representations and underscore the significance of the proposed BSA module in enhancing localization accuracy.




\begin{figure*}[tbp]
\centerline{\includegraphics[width=\textwidth]{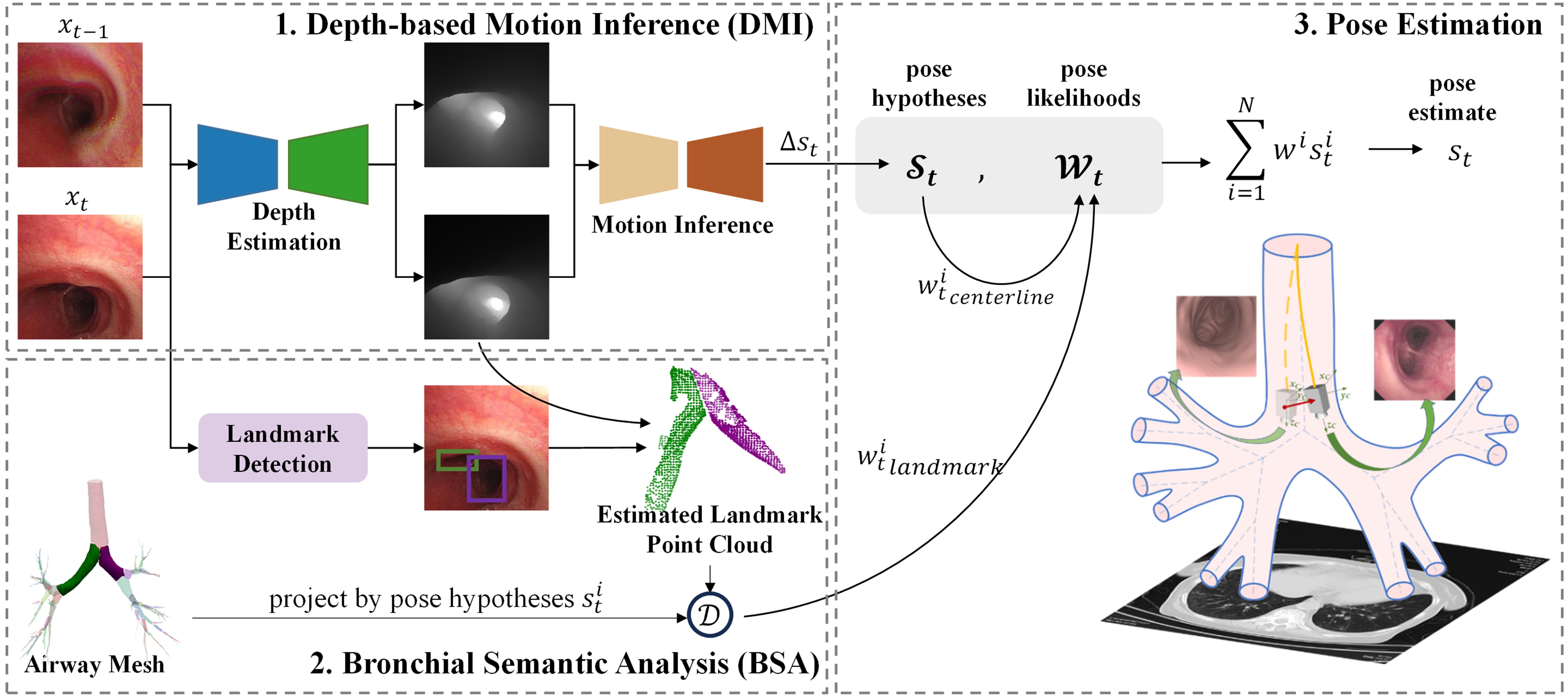}}
\caption{Overview of the proposed PANS for real-time bronchoscope localization.}
\label{fig1}
\end{figure*}

\section{Method}
\subsection{Overview}
Our PANS is illustrated in Fig. \ref{fig1}. 
We consider the problem of estimating the 6 DOF pose of the endoscope as inferring a latent state $s$ from camera image observations $x$. Due to the challenges in parameterizing the inherent uncertainties in the motion and operating conditions of the endoscope, we adopt a non-parametric way of representing probability distribution for the current endoscope pose state $s_t$, or commonly referred to as belief. This is achieved by utilizing a particle filter to approximate the belief distribution through a set of discrete samples or particles $S_t = \{s_t^{[1]}, s_t^{[2]}, \ldots, s_t^{[N]}\}$. Each particle $s_t^{[i]}$ ($1 \leq i \leq N$) is a hypothesis of the endoscope state $s_t$ at time $t$, and $N$ is the number of particles. The likelihood of each particle's represented state is denoted in the set $W_t = \{w_t^{[1]}, w_t^{[2]}, \ldots, w_t^{[N]}\}$. The essence of endoscopic tracking within our framework lies in generating $S_t$ that more closely approximates the actual endoscopic state and in accurately determining $W_t$.

The current belief $S_t$ is defined through a recursive Bayesian update. Initially, we estimate the endoscope's odometry by deriving the relative motion from $x_{t-1}$ and $x_t$, represented by $\Delta s$. We propose a Depth-based Motion Inference (DMI) module to calculate $\Delta s$ (§ 2.2). This estimate is then used to propagate the particle set $S_t$ by sampling from the state transition distribution:
\begin{equation}
    s_t^{[i]} \sim p(s_t | s_{t-1}^{[i]}, \Delta s).
\end{equation}
Subsequently, the weights $W_t$ are updated based on the current observation $x_t$, using the proposed Bronchial Semantic Analysis (BSA) module (§ 2.3) as well as the centerline prior (§ 2.3), represented as:
\begin{equation}
 w_t^{[i]} = p(s_t^{[i]} | x_t).
\end{equation}

The bronchoscopic tracking pipeline is summarized in Algorithm \ref{alg:vmf_localization}.

\subsection{Depth-based Motion Inference for Pose Hypotheses}
In our PANS, we devise the DMI to accomplish motion prediction, by sampling endoscopic state particles by a noisy state transition:
\begin{equation}
    \Delta s = f_{\theta}(x_{t-1}, x_t),
    \label{motion1}
\end{equation}
\begin{equation}
    s_t^{[i]} = s_{t-1}^{[i]} \oplus \Delta s + \epsilon^{[i]} \sim \mathcal{N},
    \label{motion2}
\end{equation}
\noindent where \( f_{\theta} \) is DMI module, \( \theta \) are the feedforward network parameters, $\oplus$ denotes pose concatenation, and \( \epsilon^{[i]} \) represents Gaussian noise. Specifically, the DMI enhances case generalization through a two-step motion inference process by first estimating depth of endoscopic frames, then deducing camera motion. This approach reduces issues from variable textures and lighting, allowing a motion inference network trained on virtual depth to adapt to real endoscopic footage.

The depth estimation network is based on a cycle generative architecture \cite{zhu2017unpaired}. Given an endoscopic frame $x_t\in X$, the depth estimation network $G_{{depth}}: X \rightarrow Z$ maps the frame into its depth space $Z$, represented by \( z_t = G_{depth}(x_t) \). Adversarial loss $L_{\text{adv}}$, cycle consistency loss $L_{\text{cyc}}$ and identity loss $L_{\text{iden}}$ are adopted to guide the network in learning domain transfer and content preservation. The LS-GAN loss \cite{mao2017least} is implemented as adversarial loss $L_{\text{adv}}$ and L1 losses are used for $L_{\text{cyc}}$ and $L_{\text{iden}}$.
To counteract scale instability and potential structural alterations in objects across different frames \cite{karaoglu2021adversarial}, view consistency loss \( L_{rec} \) and geometry consistency loss \( L_{gc} \) in \cite{tian2024ddvnb} are adopted to enhance the depth estimation network's scale perception.
Concurrently, the motion inference network is trained by minimizing the L2 norm of the discrepancy between predicted transformation and ground truth transformation, utilizing depth and endoscopic pose from virtual bronchoscopy as training data.

\subsection{Bronchial Semantic Analysis for Pose Likelihoods}
During measurement updates, particle weights are computed based on the similarity between the current observation and the supposed observation of each particle state. Previous studies in bronchoscope localization have predominantly relied on image intensity or structural similarity for matching virtual and real endoscopic frames \cite{luo2014discriminative,shen2019context,banach2021visually,gu2022vision,luo2023monocular}. However, the high degree of similarity in endoscopic frames across various airway sections would lead these methods to encounter local minima, hindering their tracking performance. 

To address the limitation of existing methods \cite{luo2014discriminative,shen2019context,banach2021visually,gu2022vision,luo2023monocular}, we propose the BSA to calculate pose likelihood based on endoscopic frame semantic analysis. Specifically, we first identify the anatomical branches (\textit{e.g.}, the right upper lobe) visible in the endoscopic view as landmarks. Subsequently, the BSA assesses the likelihood of pose hypotheses by examining the alignment between the depth estimation of each identified branch and the segmented airway model. 

To overcome the similarity in different areas of the airway, instead of directly identifying anatomical branches, BSA's landmark detection encompasses a three-stage pipeline: lumen detection, tracking, and airway association, following \cite{tian2024bronchotrack}. We first detect airway lumens in the endoscopic frame without recognizing their anatomical branch label, using the high-performance YOLOv7 \cite{wang2023yolov7} detector. We then utilize motion models and deep appearance descriptors to track these detected lumens across subsequent frames. Finally, a training-free airway association module is employed to match these tracked lumens to their corresponding anatomical branches. Through this process, we map observed lumen to its anatomical branch, and propagate their anatomical branch label temporally by tracking them in successive frames.

Following the identification of airway anatomical branches, the BSA module evaluates the likelihood of each pose hypothesis by measuring the discrepancy between the depth estimation of detected branches and the projected point cloud of these branches from the particle's state, as follows:
\begin{equation}
    w_{t, l}^{[i]}  = \sum_k D(h(o_{t_k}), \text{proj}(m_k, s_t^{[i]})), \label{weight_landmark}
\end{equation}
where \( k \) is the anatomical label of detected airway branches, \( D \) is a distance metric. The depth estimation \( h(o_{t_k}) \) of each detected branch \( k \) is compared with \( \text{proj}(m_k, s_t^{[i]}) \), which signifies the point cloud \( m_k \) of segmented airway branch \( k \) projecting from world coordinate to camera coordinate.

We adapt the Chamfer Distance into a binary count metric to evaluate the overlap between two point clouds, accounting for outliers in the depth-estimated cloud. This metric \( D \) is defined as:
\begin{equation}
    D(A, B) = \frac{1}{|A|} \left| \left\{ a \in A \mid \min_{b \in B} \left\{ d(a, b) \right\} < \rho \right\} \right|
\end{equation}
where \( d \) is the Euclidean distance, \( \rho = 3\)mm is threshold for matching, and \( | \cdot | \) signifies counting of points.

\begin{algorithm}[t]
\caption{The Pipeline of PANS for Bronchoscope Localization.}
\label{alg:vmf_localization}
\begin{algorithmic}[1]

\State \textbf{Input:} Airway segments, initial bronchoscope pose, bronchoscopic frames $\{x_t\}_{t=1}^T$
\State \textbf{Output:} Estimated pose $\hat{s}_t$ for each frame $x_t$

\State Initialize particle set $S_0$ around initial pose

\For{each time step $t = 1$ to $T$}
    \State Estimate odometry $\Delta s$ using DMI (Eq. \ref{motion1}-\ref{motion2}) based on $x_{t-1}$ and $x_t$
    \State Update $S_t$ by applying $\Delta s$ to $S_{t-1}$
    \State Calculate likelihoods $W_t$ using BSA and centerline constraint (Eq. \ref{weight_landmark}-\ref{weight})
    \State Compute estimated pose $\hat{s}_t = \sum_{i=1}^{N} w_t^i s_t^i$ (Eq. \ref{pose})
    \State Resample particles $S_t$ based on $W_t$ to focus on higher probability states
\EndFor
\end{algorithmic}
\end{algorithm}

Moreover, as bronchoscopes are typically near and oriented along the tubular's centerline, we introduce a centerline constraint as prior knowledge, as:
\begin{equation}
    w_{t,c}^{[i]} = \mathcal{N}(e | 0, \sigma_1) \cdot \mathcal{N}(\phi | 0, \sigma_2), \label{weight_centerline}
\end{equation}
\noindent where \( e \) is the distance from state \( s_t^{[i]} \) to the closest centerline, and \( \phi \) represents the angle between state \( s_t^{[i]} \) and the nearest centerline. \( \sigma_1 \) and \( \sigma_2 \) are set to \( \frac{\pi}{r} \) and \( \frac{\pi}{6} \), with \( r \) being the radius of the nearest branch.

Combining $w_{t, l}^{[i]} $ and $w_{t,c}^{[i]}$, the particle weights are updated by:
\begin{equation}
w_t^{[i]} = \frac{w_{t, l}^{[i]} \cdot w_{t,c}^{[i]}}{\left\|w_{t, l}^{[i]} \cdot w_{t,c}^{[i]}\right\|}.
\label{weight}
\end{equation}

By calculating the likelihoods of endoscopic state proposals, or particles, the estimated endoscopic pose is calculated by the weighted sum of particles as:

\begin{equation}
\hat{s}_t = \sum_{i=1}^{N} w_t^i s_t^i.
\label{pose}
\end{equation}

\begin{figure*}[tbp]
\centerline{\includegraphics[width=\textwidth]{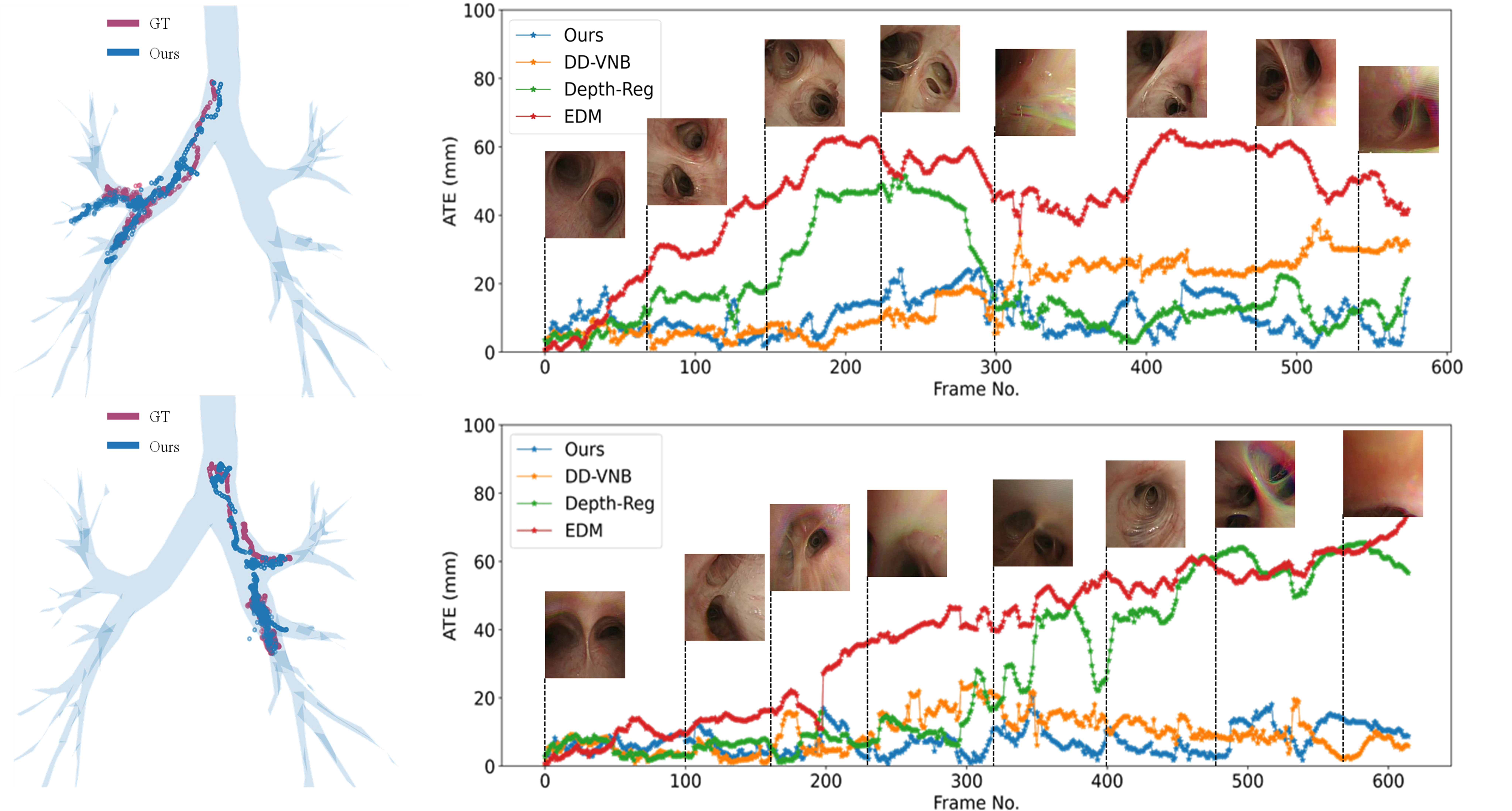}}
\caption{Localization trajectory and error on an example patient case. Two inspection trajectories into the left and right lung are presented separately.}
\label{fig2}
\end{figure*}

\section{Experiment}
\subsection{Dataset and Implementation Details}

We conducted our experiments on 31 bronchoscopic intervention cases, allocating 20 for training, 1 for validation, and 10 for testing. Each test case covers both lungs through two sequences. Bronchoscopic videos are filmed with the frame rate around 15fps using Olympus bronchoscope during regular inspection procedure, where the bronchoscope is inserted into every possible airway for thorough inspection. Patient airway meshes are semantically segmented from CT scans \cite{tan2021sgnet,xie2022structure}, categorized by branch anatomy. Our testing dataset spans shallow to deep airway branches up to the 5th generation. It features video frames with challenges like poor visibility, motion blur, and bubbles, presenting a wider range of difficulties than typically reported. Fig. \ref{fig2} displays the trajectory and select frames from a representative case. Ground truth bronchoscope poses are manually labeled by experts through registering virtual and real bronchoscopic views. More details are provided in the supplementary material.

We utilized the Pytorch framework on an NVIDIA RTX 3090 GPU for training. All networks process frames resized to 256$\times$256 pixels. The depth estimation network, based on ResNet-50 \cite{he2016deep}, was trained with a learning rate of 0.0001 over 100 epochs and a batch size of 1, adhering to the weights of losses specified in \cite{tian2024ddvnb}. The motion inference network, utilizing a FlownetC \cite{ilg2017flownet} encoder and MLP decoder, was trained for 300 epochs with a learning rate of 1e-5 and a batch size of 64. The YOLOv7 \cite{wang2023yolov7} lumen detector underwent training for 300 epochs with a batch size of 64. The appearance descriptor for lumen tracking adopts a ResNet-50 architecture trained by image retrieval follows \cite{tian2024bronchotrack}.


\begin{figure*}[tbp]
\centerline{\includegraphics[width=\textwidth]{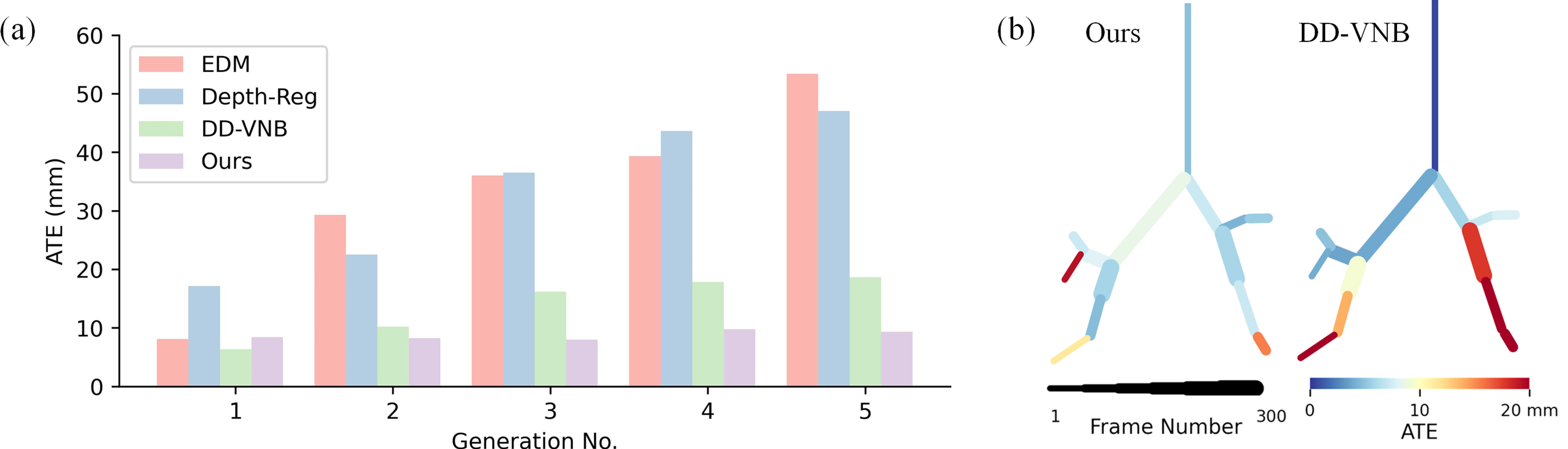}}
\caption{(a) Localization ATE on different airway generations.  (b) Localization ATE for the top two methods across various branches in an example case.}
\label{fig3}
\end{figure*}

\subsection{Comparison Results}

Our PANS is evaluated against SOTA monocular endoscopic localization techniques. This comparison encompasses depth registration (Depth-Reg) approaches \cite{shen2019context,banach2021visually}, the DD-VNB \cite{tian2024ddvnb}, and EDM \cite{recasens2021}. Evaluation metrics adhere to VNB research standards \cite{shen2019context,gu2022vision,tian2024ddvnb}. 
Table~\ref{table2} shows that our method outperforms SOTA by a significant margin across all evaluated metrics, achieving a localization ATE of \(8.7 \pm 6.0\) mm and SR10 of \(70.0\%\). When compared to methods that rely solely on registration \cite{shen2019context,banach2021visually} or visual odometry \cite{recasens2021}, PANS demonstrates superior performance, benefiting from diverse visual representations. Moreover, PANS outperforms hybrid approach \cite{tian2024ddvnb}, underscoring its effective probabilistic framework and semantic analysis. Fig. \ref{fig2} illustrates a test case visualization; additional results are in the supplementary materials.

The comparative localization accuracy of our method against SOTA across various airway generations is depicted in Fig. \ref{fig3}. Our technique demonstrates consistent accuracy even as it navigates deeper into the airway generations. In contrast, DD-VNB, Depth-Reg, and EDM exhibit a decline in performance with increasing airway depth. Specifically, EDM, which relies on incremental camera motion estimation, experiences a progressive deviation from the ground truth over time. Both DD-VNB and Depth-Reg struggle in the deeper airways, potentially due to the structural homogeneity present in the smaller bronchi.

\begin{table}[t]
    \centering
    \begin{minipage}{0.49\textwidth}
        \centering
        \caption{Localization across the 10-patient cases encompassing 20 sequences and 10,004 frames.}
        \label{table2}
        \begin{tabular}{lccc}
            \hline
            Method & ATE (mm)$\downarrow$ & SR5$\uparrow$  & SR10$\uparrow$ \\
            \hline
            EDM   & 35.7 ± 23.2 & 3.90\% & 8.8\% \\
            Depth-Reg & 35.1 ± 28.4 & 13.9\% & 25.7\% \\
            DD-VNB & 15.0 ± 11.7 & 22.9\% & 44.3\% \\
            \hline
            PANS \textit{w/o DVR} & 18.9 ± 18.6 & 15.1\% & 40.1\% \\
            PANS \textit{w/o BSA} & 15.1 ± 13.1 & 21.8\% & 49.2\% \\
            PANS  & \textbf{8.7 ± 6.0} & \textbf{28.4\%} & \textbf{70.0\%} \\\hline
        \end{tabular}
    \end{minipage}\hfill
    \hspace{0.5cm}%
    \begin{minipage}{0.40\textwidth}
        \centering
        \includegraphics[width=\linewidth]{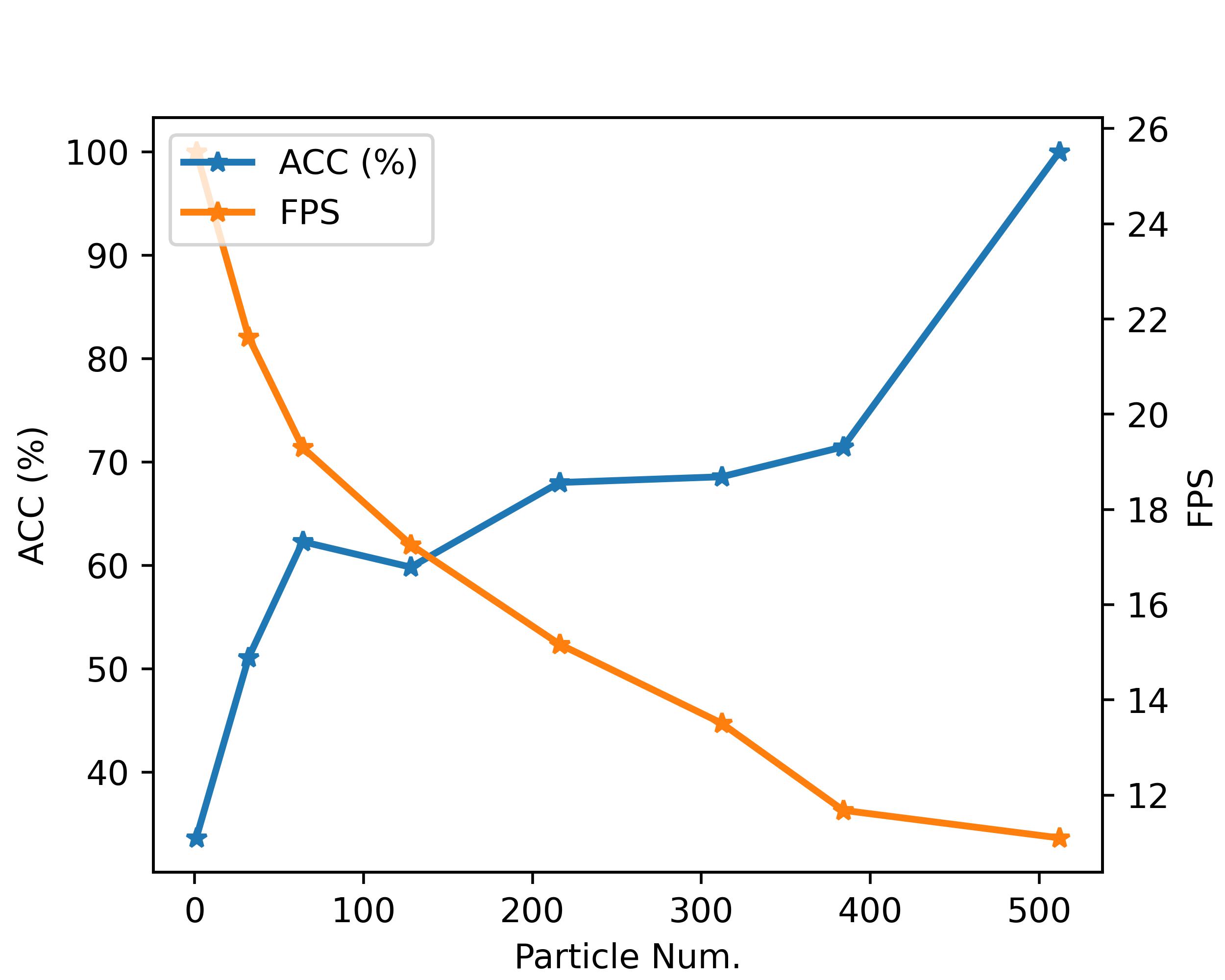}
        \captionof{figure}{Performance of PANS with different particle numbers.}
        \label{fig4}
    \end{minipage}
\end{table}

\subsection{Ablation Studies}
We assess the impact of Diverse Visual Representations (DVR) and BSA in Table \ref{table2}. The PANS \textit{w/o DVR} variant, which excludes the landmark weight and localize depending solely on motion inference with centerline guidance, demonstrates a notable drop in performance, highlighting the critical contribution of multiple visual representations to accuracy and robustness. Furthermore, PANS \textit{w/o BSA} substitutes landmark-based matching with global NCC depth matching, following the approach in \cite{tian2024ddvnb,shen2019context}. The increase in SR10 indicates enhanced tracking success across more frames, suggesting greater localization robustness.


We also conduct a particle number sensitivity test on the validation set to assess the trade-off of accuracy and computational efficiency \cite{thrun2002probabilistic}, by normalizing the lowest ATE to 100\% accuracy, as illustrated in Fig. \ref{fig4}. For comparison and ablation studies, we set the particle number to 216, aligning PANS's speed with the video capture rate of approximately 15Hz for real-time processing. Our PANS surpasses other methods in speed, especially those independent of case-specific training. Depth-Reg methods such as Shen et al. \cite{shen2019context}  operate at 0.35 fps, and Banach et al. \cite{banach2021visually} take roughly 2s per camera pose. Other approaches include Gu et al. \cite{gu2022vision} at 5 fps, Luo et al. \cite{luo2023monocular} at 0.41s per frame, and Wang et al. \cite{wang2020visual} at 0.08s per frame. Although DD-VNB \cite{tian2024ddvnb} demonstrates real-time speeds, it does not match PANS's accuracy and robustness in deeper airway generations.

\section{Conclusion}
In this work, we propose an innovative bronchoscope localization framework, named PANS, leveraging comprehensive deep representations into a Monte-Carlo framework to enhance robustness. Specifically, we propose the Depth-based Motion Inference (DMI) module to propagate endoscopic pose hypotheses overtime, and the Bronchial Semantic Analysis (BSA) module that enhances pose likelihood measurement by distinguishing airway landmarks in bronchoscopic frames. Tested on a 10-patient dataset from real interventions, our PANS excelled in tracking within complex airway branches, achieving real-time performance and generalization beyond the training data. 

%
%
%
\bibliographystyle{splncs04}
\bibliography{references}
%




\end{document}